\documentclass[10pt,twocolumn,letterpaper]{article}
\usepackage[T1]{fontenc}
\usepackage{amsmath} 
\DeclareMathOperator*{\argmin}{arg\,min} 

\usepackage{cvpr}              

\usepackage{xspace}
\newcommand{\ours}{LiveAnimate\xspace}
\DeclareMathOperator*{\argmax}{arg\,max}

\definecolor{cvprblue}{rgb}{0.21,0.49,0.74}
\usepackage[pagebackref,breaklinks,colorlinks,allcolors=cvprblue]{hyperref}
\usepackage{booktabs}  
\usepackage{pifont}    
\usepackage{caption}   
\usepackage{array}     
\usepackage{multirow}  
\usepackage{cuted}     
\usepackage{colortbl}
\usepackage{xcolor}
\definecolor{cvprblue}{RGB}{0, 82, 147}
\usepackage{arydshln}
\usepackage{graphicx}
\usepackage{lipsum}
\usepackage{indentfirst}
\usepackage{amsmath}
\usepackage{cases}
\usepackage{booktabs}
\usepackage{multirow} 
\usepackage{graphicx}
\usepackage{times}  
\usepackage{algorithm}
\usepackage{algorithmic}
\usepackage{color}
\usepackage{pifont}
\usepackage{arydshln} 
\usepackage{float} 
\usepackage{bm}
\usepackage{caption}
\usepackage{listings}
\def\paperID{*****} 
\def\confName{CVPR}
\def\confYear{2026}

\title{LiveAnimate: Stable Long-Form Streaming Human Animation in Real-Time}

\author{
Yuxuan Zhang \textsuperscript{1,2}
\quad Haozhong Xiong \textsuperscript{2}
\quad Yubo Huang \textsuperscript{2}
\quad Jiayi Song \textsuperscript{2}
\quad Jinpeng Yu \textsuperscript{2} \\
\quad Haofan Wang \textsuperscript{3} 
\quad Jiaming Liu \textsuperscript{2} 
\quad Ruihua Huang \textsuperscript{2} 
\quad Liwei Wang \textsuperscript{1}
\\
\textsuperscript{1} The Chinese University of Hong Kong, \\
\textsuperscript{2} Qwen Applications Business Group of Alibaba, \textsuperscript{3} Liblib AI \\
{\tt\small yxzhang@cse.cuhk.edu.hk, jmliu1217@gmail.com} \\
{\url{https://liveanimate.github.io/}}
}

\begin{document}
\maketitle

\begin{strip}
    \centering
    \includegraphics[width=0.9\textwidth]{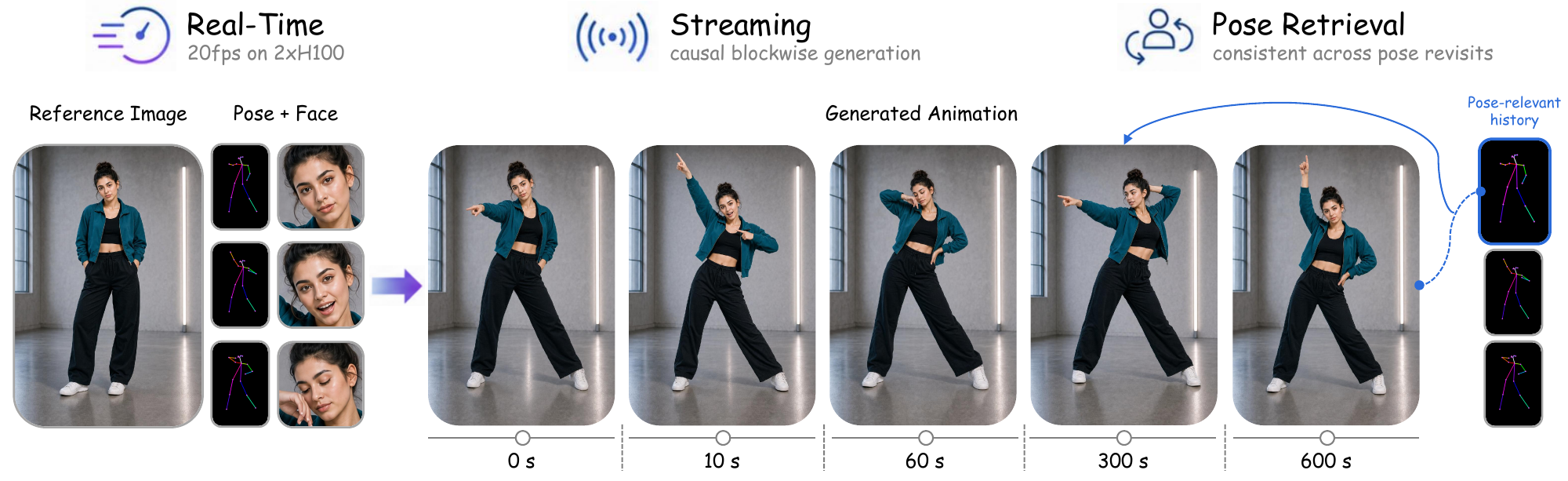}
    \captionof{figure}{\textbf{LiveAnimate enables real-time, stable long-form streaming human animation.} Given a reference image and a stream of body-pose and facial controls, LiveAnimate generates identity-consistent animation causally, one temporal block at a time. Pose-Retrieval Sink Attention recalls pose-relevant historical context when similar poses recur, preserving appearance over extended streams. With three-step sampling, LiveAnimate runs at $\sim$ 20\,FPS on two NVIDIA H100 GPUs while maintaining stable long-form generation.}
    \label{fig:teaser}
\end{strip}

\begin{abstract}
Pose-driven human animation synthesizes a video of a target person from a single reference image and a driving pose stream. Real-time generation is essential for interactive applications such as live streaming, telepresence, and virtual avatars, yet diffusion-based systems require minutes to hours per clip, precluding responsive interaction. We present \ours, to our knowledge the first animation system to combine real-time streaming with stable long-form generation at billion scale, built on a 14B-parameter video Diffusion Transformer (DiT). A two-stage training pipeline first adapts a pretrained bidirectional DiT into a block-causal autoregressive generator through Reference-Anchored Teacher-Forcing Adaptation, and then reduces the sampling budget to three steps through Block-wise Self-Forcing Distillation. To preserve appearance over extended streams, we introduce Pose-Retrieval Sink Attention (PR-Sink), a bounded KV-cache mechanism combining a Static Sink that permanently anchors the first generated block, a Dynamic Sink that holds a pose-retrieved historical block, and a three-slot Rolling Window. When a pose recurs, PR-Sink restores the relevant appearance context without retaining the entire sequence, so memory and per-block latency remain constant regardless of stream duration. Together with Ulysses sequence parallelism and operator fusion, these designs enable 19.63\,FPS streaming inference on two NVIDIA H100 GPUs. On a three-minute benchmark, \ours maintains nearly constant perceptual quality and identity from the first 30 seconds to the final minute (IQA 4.047 vs.\ 4.026), while prior systems degrade substantially or require hours of offline computation for the same rollout. These results establish a new operating point in quality, latency, and duration for interactive full-body animation.

\end{abstract}
    
\section{Introduction}
\label{sec:intro}

Pose-driven human animation, which synthesizes a video of a target person following a sequence of driving poses while preserving the appearance from a single reference image, is a fundamental task with broad applications in virtual try-on, digital human creation, live streaming, and telepresence. The core challenge is to simultaneously maintain appearance fidelity, pose accuracy, and temporal coherence over extended sequences, a combination that has proven difficult to achieve. Video diffusion models~\cite{ho2020denoising,rombach2022ldm,peebles2023dit} have become the prevailing approach to this task. Animate Anyone~\cite{hu2024animate}, MagicAnimate~\cite{xu2024magicanimate}, and UniAnimate-DiT~\cite{wang2025unianimatedit} adopt reference-based conditioning with temporal attention for short-clip generation. Building on billion-scale Diffusion Transformers (DiTs), Wan-Animate~\cite{wan2025,wananimate2025} and SCAIL~\cite{yan2025scail} scale this design to 14B-parameter backbones with 3D-consistent pose representations. More recently, EverAnimate~\cite{li2025everanimate} extends generation to the minute scale through persistent latent propagation and restorative flow matching.

\begin{table}[t]
\centering
\caption{\textbf{Capability comparison of human animation methods.} \ours is the first to achieve real-time streaming with stable long-form generation at billion scale.}
\label{tab:capability}
\vspace{-2mm}
\resizebox{\columnwidth}{!}{
\begin{tabular}{l|cccc}
\toprule
Method & Streaming & Real-Time & Stable Long-Form & Param \\
\midrule
Animate Anyone~\cite{hu2024animate}         & \ding{55} & \ding{55} & \ding{55} & -- \\
UniAnimate-DiT~\cite{wang2025unianimatedit}    & \ding{55} & \ding{55} & \ding{55} & 14B \\
One-to-All~\cite{shi2025onetoall}           & \ding{55} & \ding{55} & \ding{55} & 14B \\
SCAIL~\cite{yan2025scail}                  & \ding{55} & \ding{55} & \ding{55} & 14B \\
Wan-Animate~\cite{wan2025}                  & \ding{55} & \ding{55} & \ding{55} & 14B \\
EverAnimate~\cite{li2025everanimate}        & \ding{55} & \ding{55} & \ding{51} & 14B \\
\midrule
\rowcolor{gray!15} \textbf{\ours (Ours)}    & \ding{51} & \ding{51} & \ding{51} & \textbf{14B} \\
\bottomrule
\end{tabular}
}
\vspace{-3mm}
\end{table}

However, a critical gap remains: all of these methods are offline, requiring minutes to hours per clip and precluding responsive interaction. As shown in Table~\ref{tab:capability}, no prior full-body diffusion system simultaneously supports streaming generation, real-time inference, and stable long-form output. Most methods~\cite{hu2024animate,wang2025unianimatedit,yan2025scail} generate fixed-length clips and provide no mechanism for processing an open-ended pose stream. Even the minute-scale EverAnimate~\cite{li2025everanimate} still requires 20 denoising steps per chunk, keeping it far from real time. Closing this gap requires solving three coupled problems: \textit{(i)} reducing the inference cost of a billion-scale DiT to meet an interactive latency budget, \textit{(ii)} converting a bidirectional video model into a causal generator without sacrificing quality, and \textit{(iii)} preventing identity and appearance drift over arbitrarily long rollouts.

To address these challenges, we present \ours, the first system to enable real-time streaming human animation with stable long-form generation using a 14B-parameter causal DiT. Our approach is built on a \textit{two-stage training pipeline}. In Stage~1, \textit{Reference-Anchored Teacher-Forcing Adaptation} converts the pretrained bidirectional DiT into a block-causal generator by conditioning each training block on ground-truth clean history. A global Ref Sink makes the reference-image latent visible to all generated blocks and is distinct from the generated-history sinks used at inference. In Stage~2, \textit{Block-wise Self-Forcing Distillation} (BS-DMD), building on Self Forcing~\cite{huang2025selfforcing}, first performs a Self-Forcing Rollout without gradient tracking and then applies Block-wise DMD Optimization to one replayed block at a time. This exposes every block position to a distribution-matching signal without retaining the full rollout graph, allowing Stage-2 distillation of the 14B model to run on a single 8$\times$80GB GPU node. To maintain temporal coherence over long-form streaming, we propose \textit{Pose-Retrieval Sink Attention (PR-Sink)}, which combines a Static Sink, a pose-retrieved Dynamic Sink, and a bounded Rolling Window. The bank update objective maximizes pose coverage, while retrieval supplies the historical block written into the Dynamic Sink. At the infrastructure level, we employ Ulysses-style sequence parallelism~\cite{fang2024usp} and operator fusion to distribute attention computation across GPUs, achieving 19.63\,FPS real-time streaming on 2$\times$H100 GPUs.

In summary, our contributions are threefold:
\begin{itemize}
    \item We present \ours, the first system to enable real-time streaming human animation with stable long-form generation at billion scale, achieving $\sim$20\,FPS on 2$\times$H100 GPUs.
    \item We design a two-stage training pipeline: Reference-Anchored Teacher-Forcing Adaptation converts a pretrained bidirectional DiT into a block-causal generator, and Block-wise Self-Forcing Distillation reduces the sampling budget to three steps, with a one-block-at-a-time replay scheme that keeps training feasible on a single 8$\times$80GB GPU node. We further propose PR-Sink, a bounded KV-cache strategy that combines a Static Sink, a pose-retrieved Dynamic Sink, and a Rolling Window to maintain appearance consistency over long-form streams.
    \item On a three-minute long-form benchmark, \ours preserves perceptual quality and identity across the rollout while producing video at $\sim$ 20\,FPS, establishing a favorable trade-off among quality, latency, and duration against state-of-the-art offline systems.
\end{itemize}

\section{Related Work}
\label{sec:related}

\subsection{Video Diffusion Models}

Video diffusion models have evolved from pixel-space U-Nets~\cite{ho2020denoising,ho2022videoddpm} to latent-space architectures and Diffusion Transformers (DiTs)~\cite{peebles2023dit}. Video LDM~\cite{blattmann2023lvdm} and Stable Video Diffusion~\cite{blattmann2023svd} showed that operating in a compressed VAE latent space enables high-resolution video generation at manageable cost. More recently, billion-scale DiTs have become strong backbones for controllable generation; Wan~\cite{wan2025}, for example, scales to 14B parameters with a causal 3D VAE and a flow-matching objective. Our work builds on Wan2.2-Animate-14B\cite{wananimate2025} and adapts it from offline clip generation to causal, real-time streaming.

\subsection{Pose-Driven Human Animation}

Pose-driven human animation synthesizes realistic video of a target person performing motions specified by a driving signal such as skeleton sequences or DensePose maps. Early methods based on GANs with warping-based generators~\cite{siarohin2019fomm,wang2018vid2vid} achieve real-time speed but produce artifacts under large pose changes.

Diffusion models have since become the prevailing approach to this task. Animate Anyone~\cite{hu2024animate} introduced ReferenceNet for appearance encoding together with a pose guider and temporal attention, while MagicAnimate~\cite{xu2024magicanimate} used DensePose conditioning and segment-wise generation. MusePose~\cite{tong2024musepose} released an open-source pose-driven image-to-video framework following a similar reference-and-pose conditioning paradigm. Champ~\cite{zhu2024champ} instead employed 3D parametric body guidance to improve shape and motion controllability. UniAnimate-DiT~\cite{wang2025unianimatedit} unified reference attention with long video generation using a DiT backbone. HumanDiT~\cite{gan2025humandit} likewise studied long-form pose-guided human video generation with a DiT backbone. EverAnimate~\cite{li2025everanimate} further extends generation to the minute scale, mitigating long-horizon identity drift through persistent latent propagation and restorative flow matching. Some systems further extend controllability and visual fidelity. One-to-All~\cite{shi2025onetoall} proposed alignment-free character animation and image pose transfer; SteadyDancer~\cite{zhang2025steadydancer} emphasized first-frame preservation; and SCAIL~\cite{yan2025scail} introduced 3D-consistent pose representations. Wan-Animate~\cite{wan2025,wananimate2025} unified character animation and replacement on a large-scale video DiT. MultiAnimate~\cite{hu2026multianimate} extended pose-guided animation to multiple characters, while VACE~\cite{jiang2025vace} provided a general framework for controllable video creation and editing. Most of these methods are designed for offline generation and do not jointly target low-latency streaming and minute-scale temporal stability.

\subsection{Autoregressive Video Generation}

Autoregressive chunk-wise generation extends video beyond a model's training horizon, but repeated rollouts can accumulate visual degradation and semantic drift. Diffusion Forcing~\cite{chen2024diffusionforcing} introduced per-token noise levels for sequence generation, and CausVid~\cite{yin2025causvid} distilled a bidirectional video diffusion model into a fast causal generator. Self Forcing~\cite{huang2025selfforcing}, building on distribution matching distillation~\cite{yin2024dmd}, trained on the model's own autoregressive rollouts to reduce exposure bias. Causal Forcing~\cite{zhu2026causalforcing} argued that distilling an autoregressive student directly from a bidirectional teacher violates frame-level ODE injectivity, and instead performs distillation from a teacher-forced autoregressive model. Rolling Forcing~\cite{liu2026rollingforcing} further aligned rolling-window training with real-time long-video inference, while Context Forcing~\cite{chen2026contextforcing} used long-context conditioning for consistent autoregressive generation. Stable Video Infinity~\cite{li2026svi} recycles accumulated errors, and Rolling Sink~\cite{li2026rollingsink} dynamically refreshes attention-sink frames to bridge limited-horizon training and open-ended testing. Collectively, these methods mitigate rollout drift in general video generation, but they do not address pose-aware cache retention under strict real-time latency budgets.



\section{Method}
\label{sec:method}

\subsection{Overview}

\begin{figure*}[t]
  \centering
  \includegraphics[width=\linewidth]{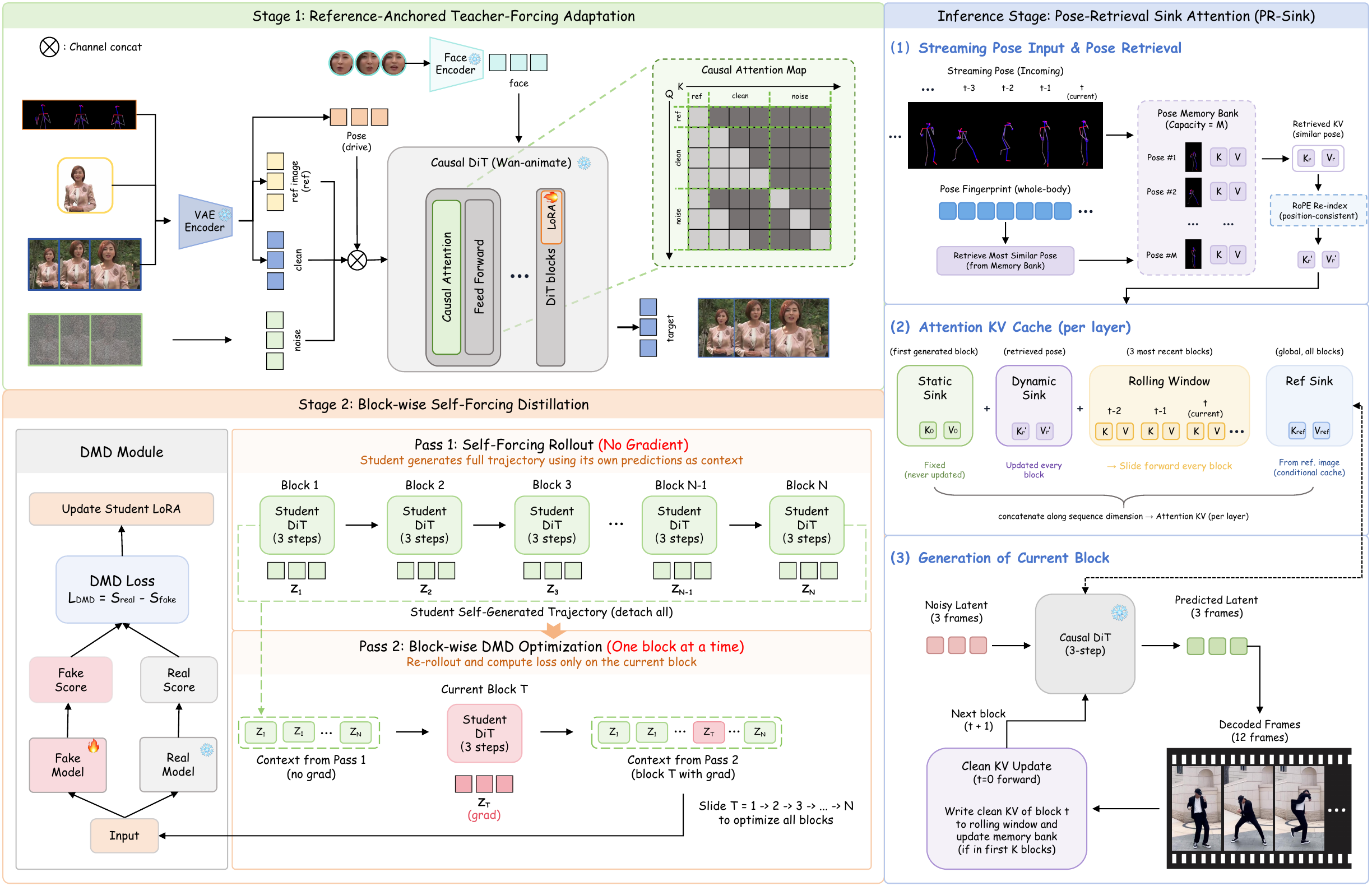}
  \caption{\textbf{Overview of \ours.} Given a reference image and streaming pose signals, our system generates video blocks autoregressively. Each block undergoes 3-step denoising followed by a clean KV update. PR-Sink augments a three-block rolling window with the first generated block and a pose-matched historical block selected from a compact memory bank. Ulysses sequence parallelism distributes attention computation across GPUs.}
  \label{fig:pipeline}
\end{figure*}

In this section, we present \ours, a framework for real-time streaming pose-driven human animation using a billion-scale causal Diffusion Transformer. As illustrated in Figure~\ref{fig:pipeline}, given a reference image $I_{\text{ref}}$ and a stream of driving pose skeletons $\{P_t\}$, our system generates video blocks autoregressively, each denoised in 3 steps and followed by a clean KV cache update. The framework consists of three components: (1) a two-stage training pipeline comprising Reference-Anchored Teacher-Forcing Adaptation (Sec.~\ref{sec:stage1}) and Block-wise Self-Forcing Distillation (Sec.~\ref{sec:stage2}); (2) Pose-Retrieval Sink Attention (PR-Sink, Sec.~\ref{sec:pr_sink}), a bounded KV-cache mechanism for stable long-form streaming; and (3) System Optimizations that enable 19.63\,FPS inference on 2$\times$H100 GPUs (Sec.~\ref{sec:optimization}).

\subsection{Stage 1: Reference-Anchored Teacher-Forcing Adaptation}
\label{sec:stage1}

Standard video diffusion models denoise complete sequences with bidirectional attention and therefore cannot directly emit temporal blocks causally. We adapt the pretrained DiT into a block-causal autoregressive generator through Reference-Anchored Teacher-Forcing Adaptation.

We partition the training video into temporal blocks of $B_f$ latent frames. During Stage~1, each block $b$ is denoised while attending to ground-truth clean target blocks $z_{<b}^{\mathrm{GT}}$ rather than to its own previous predictions. This ground-truth clean history prevents autoregressive errors from entering the conditioning context during Stage~1; exposure to student-generated history is introduced only in Stage~2.

A key distinction from standard teacher forcing~\cite{williams1989teacherforcing} in unconditional generation is the \textit{Ref Sink}, a global conditional-cache region that stores the reference-image KV states. In the conditional pose-driven setting, the model must maintain identity and appearance consistency with a given reference image throughout the entire generation. We encode the reference latent $z_{\text{ref}}$ through a forward pass at the clean timestep ($t{=}0$) and make the resulting Ref Sink globally visible to all subsequent temporal blocks. For target-block queries $Q_b$, block-causal self-attention reads the Ref Sink, the ground-truth clean history, and the current block:

\begin{equation}
    \text{Attn}(Q_b, K, V) = \text{softmax}\!\left(\frac{Q_b [K_{\text{ref}}; K_{<b}]^T}{\sqrt{d}}\right) [V_{\text{ref}}; V_{<b}]
\end{equation}
where $K_{\text{ref}}, V_{\text{ref}}$ are the key-value pairs from the reference frame, and $K_{<b}, V_{<b}$ are the teacher-provided clean KV states from blocks preceding $b$. This design ensures the reference frame serves as a permanent context anchor---analogous to attention sinks in LLMs~\cite{xiao2024streaming}---while adapting teacher forcing from text-to-video to the conditional generation setting.

At causal inference and during the Self-Forcing Rollout in Stage~2, a denoised block $\hat z_b^0$ is written to the historical cache through the \textit{Clean KV Update}, which performs an additional forward pass at the clean timestep:
\begin{equation}
    \text{KV}_{\text{cache}} \leftarrow \text{KV}_{\text{cache}} \cup \text{Forward}(\hat{z}_b^0, t{=}0, \mathcal{C})
\end{equation}
The Clean KV Update places historical states at the clean noise level used by the causal context. It does not remove errors already present in the generated latent.

\subsection{Stage 2: Block-wise Self-Forcing Distillation}
\label{sec:stage2}

The teacher-forcing adaptation produces a high-quality causal generator, but it still requires 50 denoising steps per block, precluding real-time inference. In Stage~2, we reduce the sampling budget to three steps through \textit{block-wise self-forcing distribution matching distillation} (BS-DMD), building on Self Forcing~\cite{huang2025selfforcing}. Unlike Stage~1 teacher forcing, which conditions each training block on clean ground-truth history, Self Forcing exposes the student to autoregressive histories induced by its own predictions and thereby reduces the train--inference exposure gap.

Directly backpropagating through an $N$-block self-forcing rollout would require retaining the activation graph of the entire autoregressive trajectory. We instead use a two-pass block-wise replay procedure. In the first pass, the student generates a complete trajectory without gradient tracking:
\begin{equation}
    (\bar z_1,\ldots,\bar z_N)
    = \operatorname{Rollout}_{\theta}(\epsilon_{1:N},\mathcal C)
\end{equation}
Every block is therefore generated under student-induced history, while the rollout activations are not retained for backward.

In the second pass, we replay one block position $T$ at a time. The contextual latent and KV states reconstructed from the first-pass trajectory are detached, and only the current block is recomputed with gradient tracking:
\begin{equation}
    z_T^{\theta}=G_{\theta}\!\left(
    \epsilon_T;
    \operatorname{sg}\!\left(\operatorname{KV}(\bar z_{<T})\right),
    \mathcal C_T\right),
\end{equation}
where $\operatorname{sg}(\cdot)$ denotes stop-gradient. We insert the recomputed block into the otherwise detached trajectory,
\begin{equation}
    \tilde z^{(T)}=
    [\operatorname{sg}(\bar z_1),\ldots,z_T^{\theta},\ldots,
    \operatorname{sg}(\bar z_N)],
\end{equation}
and evaluate the distribution-matching objective on the concatenated trajectory:
\begin{equation}
    \mathcal L_T=\mathcal L_{\mathrm{DMD}}\!\left(
    \tilde z^{(T)};\mathcal C\right),
    \qquad T=1,\ldots,N.
\end{equation}
The teacher and critic thus evaluate a complete student-induced trajectory, whereas gradients reach the student only through the currently replayed block $z_T^{\theta}$. We backpropagate $\mathcal L_T$ before proceeding to the next block and release the current block's activation graph. Sliding $T$ from 1 to $N$ provides a direct training signal to every block position without retaining backward activations for all blocks simultaneously. Together with LoRA adaptation, this keeps peak memory low enough to distill the 14B-parameter student on a single 8$\times$80GB GPU node.

\subsection{Pose-Retrieval Sink Attention}
\label{sec:pr_sink}

Autoregressive streaming presents a memory--context dilemma. Retaining every historical key--value (KV) state gives complete context but makes memory and attention cost grow linearly with duration. A bounded cache has constant cost, but irreversibly discards appearance evidence once its block leaves the Rolling Window. This is especially harmful for articulated motion: when a subject returns to a previously observed pose, the Rolling Window may contain only the intervening views. Pose-Retrieval Sink Attention (PR-Sink) addresses this dilemma by combining two complementary persistent regions: a \emph{Static Sink} that permanently anchors the sequence to its first generated block, and a \emph{Dynamic Sink} whose content is refreshed with a pose-matched historical block retrieved from a compact memory bank. The Static Sink provides time-invariant appearance context, whereas the Dynamic Sink restores view- and articulation-specific evidence when a pose recurs.

\noindent\textbf{Bounded cache organization.}
For the current block $t$, each attention layer uses a Static Sink containing the first generated block, a Dynamic Sink containing a pose-retrieved historical block, and a three-slot Rolling Window containing two recent clean blocks $t{-}2$ and $t{-}1$ together with the current block $t$. Reference-image tokens are stored separately in the global Ref Sink and are not counted as either generated-history sink or as one of the Rolling Window slots. After block $t$ is denoised, the Clean KV Update performs an additional forward pass at $t{=}0$, writes the resulting KV state into the Rolling Window, and advances the window before the next block is generated. Because the Static Sink, Dynamic Sink, Rolling Window, Ref Sink, and pose bank all have fixed capacities, cache storage and attention cost do not grow with stream duration.

\noindent\textbf{Confidence-aware whole-body fingerprint.}
We extract 133 whole-body keypoints with ViTPose and convert them into the animation representation. Each frame contains body landmarks and landmarks for each hand. 
Let $\mathbf{p}_{b,t,k}=(x,y,c)$ contain the normalized coordinates and confidence of keypoint $k$ in frame $t$ of block $b$. We concatenate all landmarks from the three ordered pose frames and apply $\ell_2$ normalization:
\begin{equation}
 \boldsymbol{\phi}_b =
 \operatorname{Norm}\!\left(
 \operatorname{Concat}_{t=1}^{3}
 \{\mathbf{p}_{b,t,k}\}_{k\in\mathcal{K}}\right).
\label{eq:pose_fingerprint}
\end{equation}
Here $|\mathcal{K}_{\mathrm{body}}|=20$ and
$|\mathcal{K}_{\mathrm{lh}}|=|\mathcal{K}_{\mathrm{rh}}|=21$, yielding a
$3\times62\times3=558$ dimensional fingerprint. Temporal concatenation preserves short-term motion configuration that would be lost under mean pooling. If pose detection fails for an entire frame, its portion of the fingerprint is zero padded; otherwise confidence is retained as the third coordinate. Global $\ell_2$ normalization makes the dot product between two fingerprints equal to their cosine similarity.
\begin{figure*}[!ht]
  \centering
  \includegraphics[height=0.7\textheight]{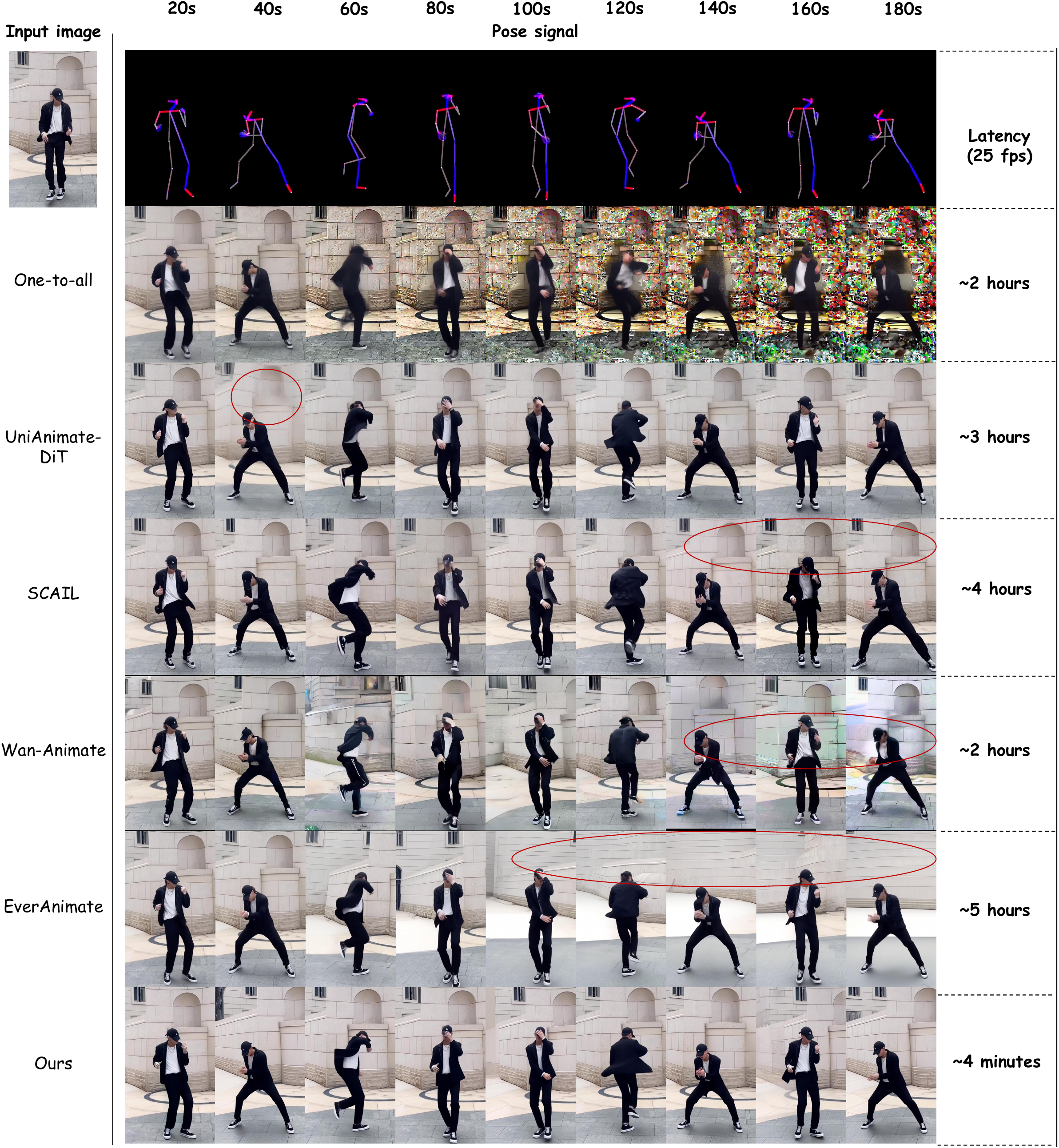}
  \caption{\textbf{Qualitative comparison on a full-body sequence.} Frames are sampled every 20 seconds from a three-minute, 25-FPS rollout. The pose signal is shown above the generated frames, and red annotations highlight representative long-horizon artifacts in competing methods. The rightmost column reports end-to-end generation time: the baselines require approximately 2--5 hours, whereas \ours completes the sequence in approximately 4 minutes.}
  \label{fig:qualitative_full}
\end{figure*}

\noindent\textbf{Diversity-preserving bank update.}
The memory bank $\mathcal{B}=\{(\boldsymbol{\phi}_i,\mathrm{KV}_i)\}$ has a fixed capacity $M=5$. Each entry stores the fingerprint and the unrotated keys and values from all transformer layers. Once the bank is full, an incoming block $q$ may replace one existing entry. We test every possible replacement and choose the bank with the lowest average pairwise cosine similarity:

\begin{equation}
 i^*=\argmin_i\;
 \operatorname{AvgSim}\!\left(
 (\mathcal{B}\setminus\{i\})\cup\{q\}\right).
\label{eq:bank_update}
\end{equation}
We perform the replacement only if it lowers the current bank similarity. Thus, writing favors \emph{coverage}: redundant poses are rejected, while novel articulations increase the diversity of the stored context. Bank updates are restricted to the first 20 blocks, after which the selected representatives are reused throughout the stream.
\begin{figure*}[!ht]
  \centering
  \includegraphics[width=\textwidth]{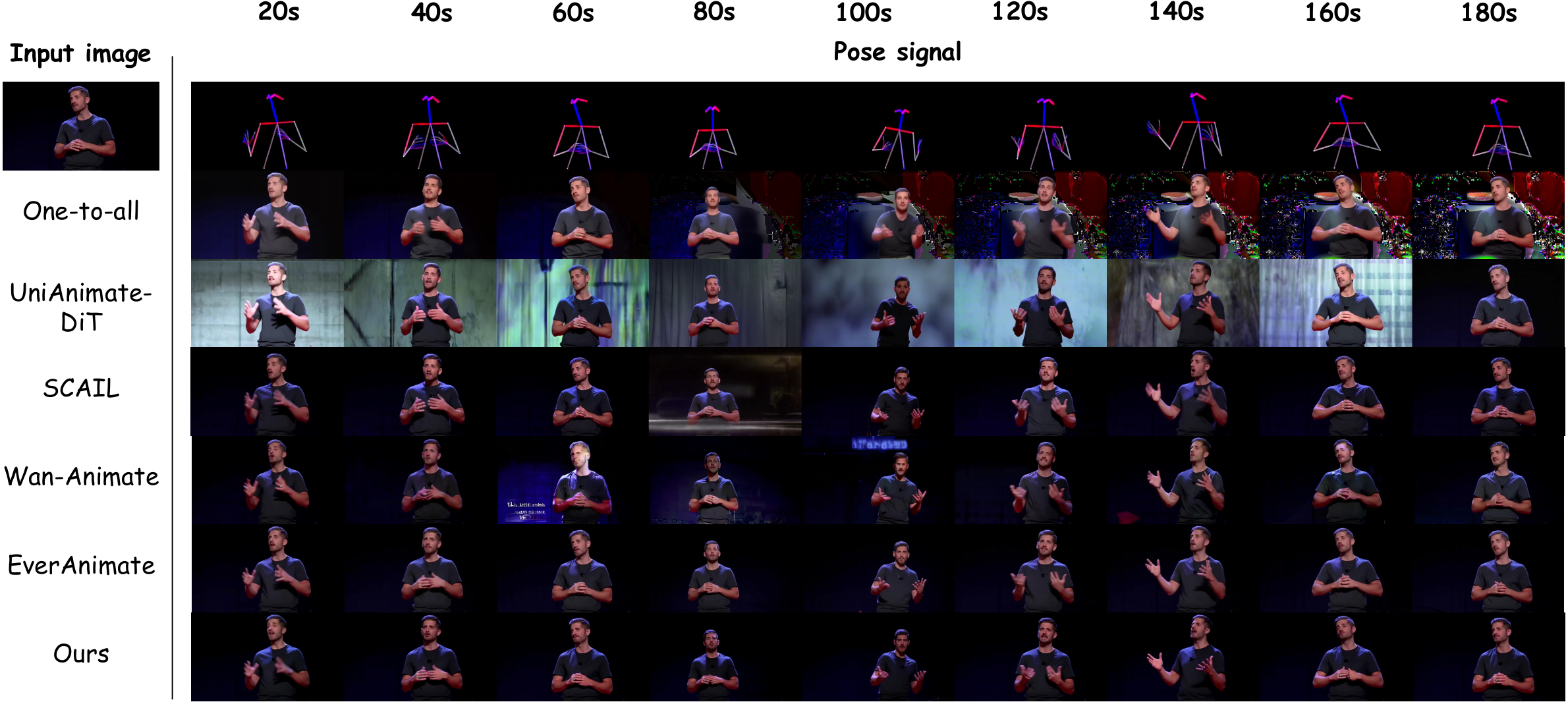}
  \caption{\textbf{Qualitative comparison on an upper-body sequence.} Frames are sampled every 20 seconds from a three-minute rollout under the same reference image and driving-pose sequence. \ours maintains the subject's identity, clothing, and dark background more consistently over time.}
  \label{fig:qualitative_upper}
\end{figure*}

\noindent\textbf{Pose-nearest retrieval.}
Before generating block $b$, we compute its fingerprint from the incoming pose signal and retrieve
\begin{equation}
 r(b)=\argmax_{i\in\mathcal{B},\,i\neq b-1}
 \boldsymbol{\phi}_b^\top\boldsymbol{\phi}_i .
\label{eq:bank_retrieval}
\end{equation}
We exclude block $b-1$ because it is already present in the Rolling Window; selecting it again would waste the Dynamic Sink on redundant local context. The retrieved KV state is copied into the Dynamic Sink. Bank writing therefore seeks \emph{coverage}, whereas bank reading seeks \emph{relevance} to the current pose. PR-Sink is the union of the pose-conditioned Dynamic Sink and the permanent first-block Static Sink; neither component alone constitutes the full mechanism.

\noindent\textbf{Position-consistent KV reuse.}
Historical blocks cannot be copied with their original rotary position embeddings (RoPE), because the Dynamic Sink has a different temporal position at retrieval time. We therefore cache $K^{\mathrm{raw}}$ before RoPE together with $V$. At attention time, keys from the Static Sink, Dynamic Sink, and Rolling Window are separately re-rotated at their assigned positions in the current cache layout before being concatenated with the current block. This decouples content selection from its original timestamp and permits the same bank entry to be reused at any later point without positional mismatch.

\subsection{System Optimizations}
\label{sec:optimization}

To meet the latency requirements of real-time streaming, we employ several infrastructure-level optimizations.

\noindent\textbf{Ulysses sequence parallelism.} We distribute computation across $N$ GPUs using the Ulysses all-to-all pattern~\cite{fang2024usp}: each GPU computes Q, K, V projections on its local token shard, then all-to-all communication redistributes to head-parallel layout where each GPU computes full-sequence attention for $H/N$ heads, followed by a reverse all-to-all gather. This requires 2 all-to-all operations per layer but each involves only head-dimension-sized tensors. The FFN computation remains embarrassingly parallel on local token shards, and the KV cache is maintained independently per GPU head shard with no additional communication overhead.

\noindent\textbf{torch.compile.} All DiT blocks are compiled with \texttt{max-autotune-no-cudagraphs} mode. We pre-compile for three resolution buckets ($480{\times}480$, $672{\times}384$, $384{\times}672$) with \texttt{dynamic=False} for maximum kernel efficiency.

\section{Experiments}
\label{sec:experiments}

\subsection{Experimental Setup}

\begin{figure*}[!t]
\centering
\includegraphics[width=\textwidth]{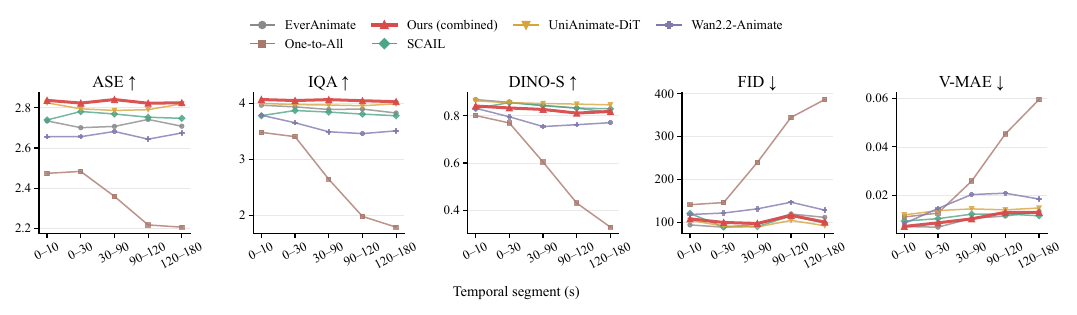}
\caption{\textbf{Quality and identity over a three-minute rollout.} Metrics are computed on a 0--10\,s prefix, the 0--30\,s initial window, and three subsequent temporal segments. The full \ours model is highlighted in red. Flat trajectories indicate resistance to accumulated degradation. Higher is better for ASE, IQA, and DINO-S; lower is better for FID and V-MAE.}
\label{fig:long_quality}
\end{figure*}
\noindent\textbf{Implementation details.}
We train LiveAnimate on 40k talking videos\cite{avspeech} and 20k human motion videos\cite{tiktokdance, humanvid}. Both training and inference support three aspect-ratio buckets: $480{\times}480$,$384{\times}672$, and $672{\times}384$ pixels. During training, \ours is initialized from the Wan2.2-Animate-14B checkpoint and fine-tuned with LoRA of rank 128. Training proceeds in two stages on 8 NVIDIA H100 GPUs. In the first stage, we train for 10,000 steps with a learning rate of $5{\times}10^{-5}$. In the second stage, we continue training for 20,000 steps with identical LoRA parameters, using learning rates of $1{\times}10^{-5}$ for both the generator and the critic. Unless stated otherwise, we generate at resolution using blocks of three latent frames (12 RGB frames). Inference is performed on two NVIDIA H100 80GB GPUs. 

\noindent\textbf{Evaluation protocol.}
We construct a three-minute benchmark consisting of 24 pairs of reference images and driving videos, divided into two complementary parts. The first part contains 12 in-the-wild videos collected from the web, covering full-body and upper-body motion, diverse appearances, and varied backgrounds. The second part supports controlled long-horizon evaluation: since long-form full-body motion footage is scarce, we derive 12 sequences from X-Dance~\cite{zhang2025steadydancer} by playing each driving pose sequence forward, in reverse, and forward again to reach three minutes, with the reference image taken from the source video. Every frame therefore remains real human motion, while each pose recurs multiple times at large temporal intervals, directly stressing identity and appearance consistency under pose recurrence. Each method receives the same reference image and driving pose sequence. We evaluate a short 0--10\,s prefix, the broader 0--30\,s initial window, and three non-overlapping later segments: 30--90, 90--120, and 120--180\,s. The nested short windows characterize initial quality, while the later segments isolate accumulated degradation.

\noindent\textbf{Metrics.}
We assess complementary aspects of generation quality without relying on pixel-aligned reconstruction metrics, which can penalize perceptually valid motion and appearance variations. Aesthetic score (ASE)\cite{qalign} and no-reference image-quality assessment (IQA)\cite{qalign} measure frame-level visual quality. DINO similarity (DINO-S)\cite{dino} measures appearance consistency with the reference identity, while FID~\cite{heusel2017fid} and VideoMAE feature distance (V-MAE)~\cite{tong2022videomae} evaluate distributional quality and temporal representation error, respectively. Higher is better for ASE, IQA, and DINO-S; lower is better for FID and V-MAE.

\noindent\textbf{Latency and throughput measurement.} 
All latency and throughput numbers reported in this paper measure the DiT generation loop only, rather than end-to-end wall-clock time. Condition processing (\eg, reference and pose encoding) and VAE encoding/decoding are excluded from the measurement, as their costs are either one-time or block-independent, can be overlapped on separate devices, and vary considerably across implementations. Since DiT forward passes dominate the recurrent per-block cost (Table~\ref{tab:latency_breakdown}), this protocol isolates the primary bottleneck and provides a fair basis for comparison across systems.

\noindent\textbf{Baselines.}
We compare with five recent full-body animation systems: EverAnimate~\cite{li2025everanimate}, One-to-All~\cite{shi2025onetoall}, SCAIL~\cite{yan2025scail}, UniAnimate-DiT~\cite{wang2025unianimatedit}, and Wan2.2-Animate~\cite{wan2025,wananimate2025}. We additionally ablate the two components of PR-Sink. The \emph{w/o Dynamic Sink} variant retains the Static Sink but disables pose-based historical retrieval, whereas the \emph{w/o Static Sink} variant retains the pose-retrieved Dynamic Sink but removes the permanent first-block anchor. The full model combines both sinks with the three-slot Rolling Window.

\noindent\textbf{Long-video baseline protocol.}
The official implementations of SCAIL and UniAnimate-DiT operate on fixed-length clips using the temporal context available during training and do not provide native long-form generation. For a controlled long-sequence comparison, we extend both methods with the same training-free sliding-window denoising procedure. We initialize a single noise latent spanning the complete target sequence. At every denoising step, we slide each model's native temporal context window across this latent with a fixed stride. Each window is conditioned on the same reference identity image and the temporally aligned local segment of the driving pose sequence. Predictions in overlapping regions are fused using triangular, or tent-shaped, weights that emphasize the temporal center of each window and suppress its boundaries. A single UniPC multistep ODE-solver update~\cite{zhao2023unipc} is then applied to the fused full-sequence prediction, ensuring that the solver state evolves consistently along one global trajectory rather than independently within each window. This extension requires neither retraining nor positional-embedding extrapolation and is used uniformly for SCAIL and UniAnimate-DiT. EverAnimate and One-to-All are instead evaluated using the long-video generation procedures provided by their official implementations.

\subsection{Quantitative Comparison}

\noindent\textbf{Short-horizon quality.}
Figure~\ref{fig:long_quality} reports quality and identity metrics across the temporal segments of the three-minute rollout. On the initial 0--30\,s segment, \ours obtains the best ASE (2.823) and IQA (4.047). This result indicates that three-step distillation preserves strong perceptual quality under a real-time inference budget.

\noindent\textbf{Long-horizon stability.}
The main advantage of \ours is its behavior as the rollout grows. Its IQA remains essentially unchanged, from 4.047 in the first segment to 4.026 in the final segment, and ASE remains stable from 2.823 to 2.824. DINO-S decreases by only 0.015, from 0.833 to 0.818. In contrast, One-to-All exhibits pronounced accumulation: IQA falls from 3.402 to 1.786, DINO-S from 0.769 to 0.328, and FID rises from 146.29 to 386.33. Wan2.2-Animate also loses identity similarity over time, with DINO-S decreasing from 0.795 to 0.770. These trends support our central claim that explicitly managing long-range context is important for streaming animation.

\noindent\textbf{Comparison with offline long-video methods.}
SCAIL and UniAnimate-DiT attain higher identity similarity (DINO-S) than \ours throughout the rollout. As the qualitative comparison shows, however, both methods exhibit inter-frame flickering (Figure~\ref{fig:qualitative_full}), whereas \ours remains stable over time. EverAnimate maintains competitive per-frame quality but is not a real-time system, requiring hours of offline computation for each three-minute sequence. \ours is thus the only evaluated method that sustains visibly consistent quality over the full rollout while generating in real time.

\begin{table}[t]
\centering
\caption{\textbf{Scaling efficiency of Ulysses sequence parallelism on H100 GPUs at $480{\times}480$.} Results are measured on the three-minute benchmark. Latency denotes the time required to generate one 12-frame block. Speedup is relative to one GPU, and efficiency is speedup divided by the number of GPUs.}
\label{tab:ulysses_scaling}
\footnotesize
\begin{tabular}{@{}rrrrr@{}}
\toprule
GPUs & Latency (ms) & FPS & Speedup & Efficiency \\
\midrule
1 & 966.96 & 12.41 & 1.00$\times$ & 100.0\% \\
2 & 611.28 & 19.63 & 1.58$\times$ & 79.1\% \\
4 & 542.25 & 22.13 & 1.78$\times$ & 44.6\% \\
\bottomrule
\end{tabular}
\end{table}

\noindent\textbf{Sequence-parallel scaling.}
Table~\ref{tab:ulysses_scaling} evaluates Ulysses scaling on the same three-minute benchmark. Moving from one to two H100 GPUs increases throughput from 12.41 to 19.63\,FPS, yielding a $1.58{\times}$ speedup at 79.1\% parallel efficiency. Expanding to four GPUs improves throughput only to 22.13\,FPS: the speedup saturates at $1.78{\times}$ and efficiency falls to 44.6\%, as the additional all-to-all communication increasingly offsets the reduction in per-GPU attention computation. We therefore adopt two-way sequence parallelism by default, as it offers the best trade-off between throughput and GPU count.

\begin{table}[t]
\centering
\caption{\textbf{Steady-state inference profile on the three-minute benchmark.} Steady state denotes inference after compilation and warm-up. Throughput is computed from 12 RGB frames per block. Component times exclude VAE decoding.}
\label{tab:latency_breakdown}
\resizebox{\columnwidth}{!}{%
\begin{tabular}{lrrrrr}
\toprule
Component & Avg. (s) & Min. (s) & Max. (s) & Share & FPS \\
\midrule
Sink read        & 0.003303 & 0.003197 & 0.005136 & 0.540\%  & -- \\
Denoising        & 0.458884 & 0.451726 & 0.476683 & 75.070\% & -- \\
Clean KV Update   & 0.149078 & 0.146292 & 0.168945 & 24.388\% & -- \\
Sink write       & 0.000011 & 0.000009 & 0.000023 & 0.0018\% & -- \\
\midrule
Block total      & 0.611276 & 0.601543 & 0.631650 & 100.00\% & 19.63 \\
\bottomrule
\end{tabular}%
}
\end{table}

\begin{figure}[h]
\centering
\includegraphics[width=\columnwidth]{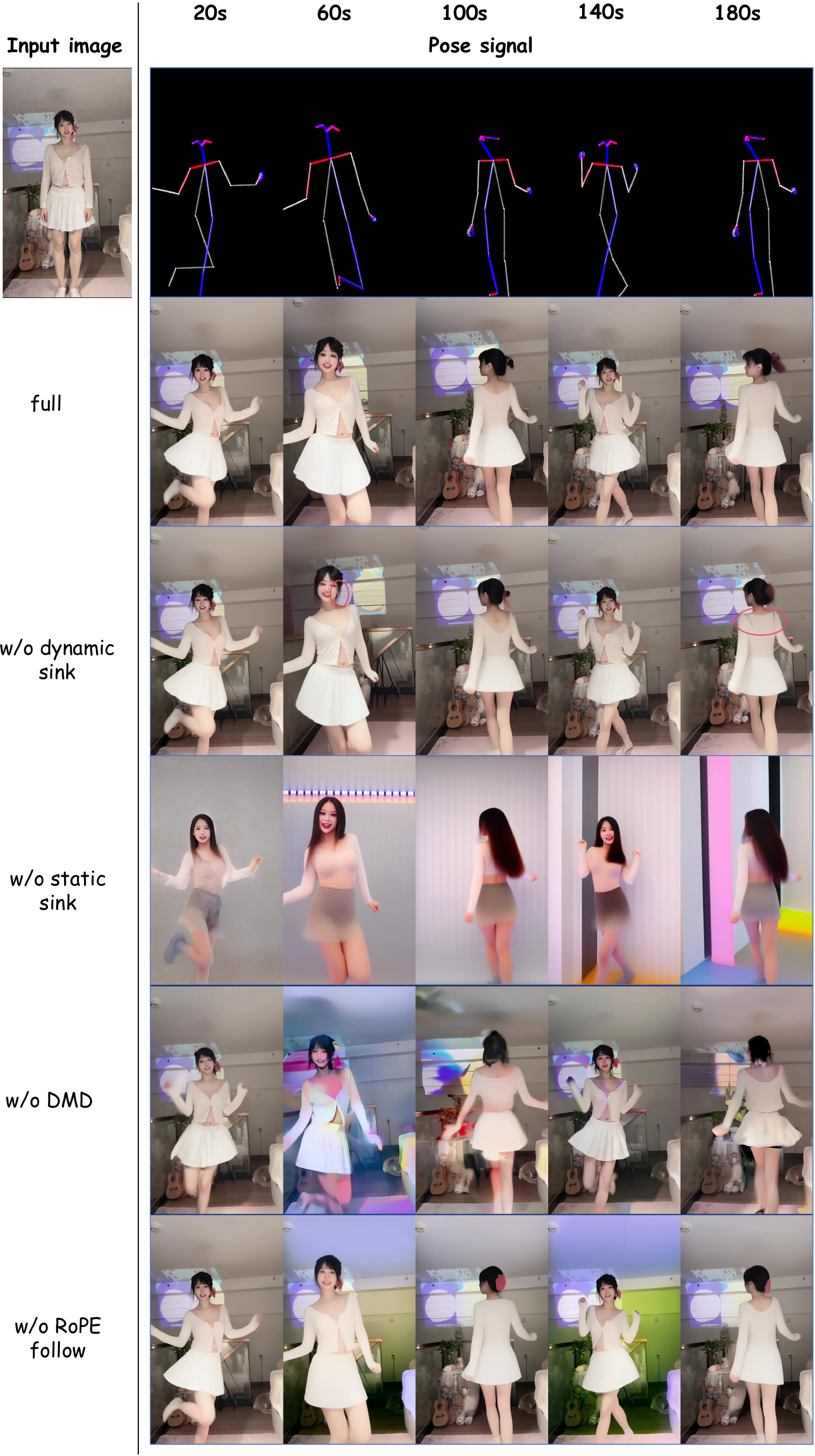}
\caption{\textbf{Qualitative ablation over a three-minute rollout.} Frames are sampled every 40 seconds from 20 to 180 seconds under the same reference image and driving-pose sequence.}
\label{fig:ablation_visual}
\end{figure}

\begin{figure*}[!t]
\centering
\includegraphics[width=\textwidth]{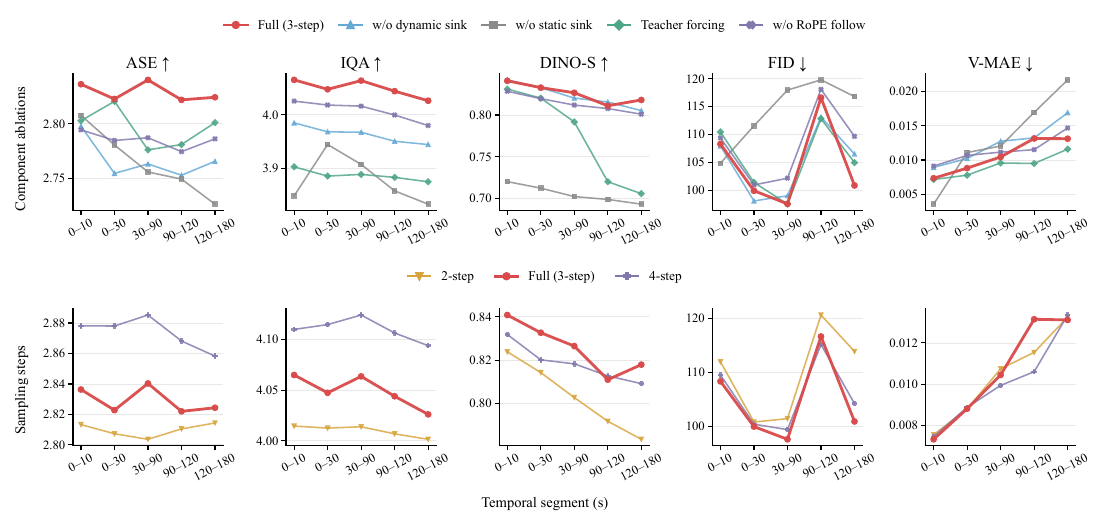}
\caption{\textbf{Quantitative ablation over the three-minute rollout.} Top: component and cache variants, including separate removal of the static and dynamic sinks. The w/o-DMD variant replaces DMD-based self-forcing distillation with conventional teacher forcing. Bottom: the sampling-step trade-off with the full PR-Sink configuration. Higher is better for ASE, IQA, and DINO-S; lower is better for FID and V-MAE.}
\label{fig:ablations}
\end{figure*}

\noindent\textbf{Inference-time breakdown.}
Table~\ref{tab:latency_breakdown} reports the steady-state latency breakdown on our three-minute benchmark. \ours processes each 12-frame block in 0.611\,s, corresponding to 19.63\,FPS. Because each block is emitted as soon as its driving poses arrive and the KV cache has a fixed size, both latency and memory are independent of the elapsed stream length. Denoising accounts for 75.07\% of the runtime, followed by the Clean KV Update at 24.39\%. The latter performs an additional forward pass at the clean timestep ($t{=}0$) and stores the resulting KV state for subsequent blocks. PR-Sink retrieval and bank maintenance together contribute only 0.542\%, demonstrating that long-range pose-aware context incurs negligible overhead. VAE decoding is excluded from the block time; it is independent across blocks and can be pipelined on a separate device concurrently with DiT inference, so it does not extend the generation critical path. Under our evaluation settings, all evaluated baselines require approximately 2--5 hours to synthesize the same three-minute sequence. \ours therefore provides a substantial speed advantage, achieving real-time streaming while maintaining bounded memory over long sequences.

\subsection{Qualitative Comparison}

Figure~\ref{fig:qualitative_full} presents the more challenging full-body example, which evaluates articulated motion, identity preservation, and background stability under large pose changes. The competing methods fail in visibly different ways. One-to-All suffers a severe collapse of image quality as the rollout progresses. UniAnimate-DiT and SCAIL maintain the global structure but exhibit inter-frame flickering. Wan-Animate and EverAnimate remain stable in the early segments but show noticeable color drift and background drift at later timestamps. In contrast, \ours follows the driving poses while preserving both the subject's appearance and the scene throughout the three minutes. The accompanying runtime comparison further highlights the efficiency gap: competing methods require approximately 2--5 hours to generate the 25-FPS sequence, whereas \ours completes it in about 4 minutes.

Figure~\ref{fig:qualitative_upper} presents an upper-body example that isolates long-term identity and appearance preservation under subtler motion. UniAnimate-DiT, SCAIL, and Wan-Animate exhibit background flickering across the sequence, and One-to-All again suffers a severe drop in image quality. EverAnimate and \ours both remain stable and consistent over the full rollout, though only \ours does so in real time.

\subsection{Ablation Study}

\noindent\textbf{Qualitative ablation.}
Figure~\ref{fig:ablation_visual} visualizes the three-minute rollout under each variant. Removing the Dynamic Sink largely preserves the scene but introduces localized appearance errors when poses recur, as highlighted around the face and shoulder. Removing the Static Sink is far more damaging: the subject and background change completely over the sequence, showing that pose-matched retrieval alone cannot provide a time-invariant identity anchor. Replacing DMD-based distillation with conventional teacher forcing causes illumination and appearance to fluctuate across the rollout, and disabling RoPE follow leads to progressive color and background shifts despite broadly correct poses. The full model remains consistent across recurring poses throughout the three minutes.

\noindent\textbf{Quantitative ablation.}
Figure~\ref{fig:ablations} reports the corresponding metric trajectories. Removing the Dynamic Sink moderately degrades long-horizon quality, reducing final-segment DINO-S from 0.818 to 0.805 and increasing FID from 100.90 to 106.52. Removing the Static Sink is substantially worse: final-segment DINO-S collapses to 0.693, yet the variant retains moderate frame-level scores (ASE 2.726, IQA 3.833), showing that frame-level metrics alone cannot diagnose identity failure. These trends confirm the complementary roles of the two sinks: the Static Sink supplies a stable global appearance anchor, while the Dynamic Sink restores pose-specific evidence that has left the Rolling Window. Disabling RoPE follow also weakens final-segment identity (DINO-S 0.801). Replacing DMD-based distillation with teacher forcing degrades the rollout steadily, with IQA falling from 3.849 to 3.438 and DINO-S from 0.780 to 0.679, indicating that self-forcing limits error accumulation under few-step sampling. Finally, four sampling steps yield the strongest frame-level scores but require an extra denoising pass, while two steps weaken long-horizon identity (final DINO-S 0.783); three steps provide the best efficiency--quality compromise.

\section{Conclusion}
\label{sec:conclusion}

We presented \ours, a streaming pose-driven human animation system that combines real-time generation with stable long-form quality. A two-stage training pipeline converts a pretrained bidirectional DiT into a block-causal generator and distills it to three sampling steps, trainable on a single 8$\times$80GB GPU node, while Pose-Retrieval Sink Attention maintains long-range appearance context under a bounded cache. On the three-minute benchmark, \ours sustains nearly constant perceptual quality and identity at 19.63\,FPS on two H100 GPUs, with memory and latency independent of stream duration, whereas offline baselines degrade visibly or require hours of computation. These results bring billion-scale video diffusion models within reach of interactive applications such as live streaming, telepresence, and virtual avatars.

\noindent\textbf{Limitations and future work.} \ours currently generates at $480{\times}480$ with three denoising steps, which limits visual fidelity and resolution, and it does not support multi-person scenes or large camera motion. Extending \ours toward higher-resolution streaming generation, multi-person animation, and camera-dynamic scenes is left to future work.

\clearpage
{
    \small
    \bibliographystyle{ieeenat_fullname}
    \bibliography{main}

\begin{thebibliography}{42}
\providecommand{\natexlab}[1]{#1}
\providecommand{\url}[1]{\texttt{#1}}
\expandafter\ifx\csname urlstyle\endcsname\relax
  \providecommand{\doi}[1]{doi: #1}\else
  \providecommand{\doi}{doi: \begingroup \urlstyle{rm}\Url}\fi

\bibitem[{Alibaba Wan Team}(2025)]{wan2025}
{Alibaba Wan Team}.
\newblock Wan: Open and advanced large-scale video generative models.
\newblock \emph{arXiv preprint arXiv:2503.20314}, 2025.

\bibitem[Blattmann et~al.(2023{\natexlab{a}})Blattmann, Dockhorn, Kulal, Mendelevitch, Kilian, Lorenz, Levi, English, Voleti, Letts, et~al.]{blattmann2023svd}
Andreas Blattmann, Tim Dockhorn, Sumith Kulal, Daniel Mendelevitch, Maciej Kilian, Dominik Lorenz, Yam Levi, Zion English, Vikram Voleti, Adam Letts, et~al.
\newblock Stable video diffusion: Scaling latent video diffusion models to large datasets.
\newblock \emph{arXiv preprint arXiv:2311.15127}, 2023{\natexlab{a}}.

\bibitem[Blattmann et~al.(2023{\natexlab{b}})Blattmann, Rombach, Ling, Dockhorn, Kim, Fidler, and Kreis]{blattmann2023lvdm}
Andreas Blattmann, Robin Rombach, Huan Ling, Tim Dockhorn, Seung~Wook Kim, Sanja Fidler, and Karsten Kreis.
\newblock Align your latents: High-resolution video synthesis with latent diffusion models.
\newblock In \emph{CVPR}, 2023{\natexlab{b}}.

\bibitem[Caron et~al.(2021)Caron, Touvron, Misra, J{\'e}gou, Mairal, Bojanowski, and Joulin]{dino}
Mathilde Caron, Hugo Touvron, Ishan Misra, Herv{\'e} J{\'e}gou, Julien Mairal, Piotr Bojanowski, and Armand Joulin.
\newblock Emerging properties in self-supervised vision transformers.
\newblock In \emph{ICCV}, 2021.

\bibitem[Chen et~al.(2024)Chen, Monso, Du, Simchowitz, Tedrake, and Sitzmann]{chen2024diffusionforcing}
Boyuan Chen, Diego~Marti Monso, Yilun Du, Max Simchowitz, Russ Tedrake, and Vincent Sitzmann.
\newblock Diffusion forcing: Next-token prediction meets full-sequence diffusion.
\newblock In \emph{NeurIPS}, 2024.

\bibitem[Chen et~al.(2026)Chen, Wei, Sun, Nie, Zhou, Zhang, Yang, and Chen]{chen2026contextforcing}
Shuo Chen, Cong Wei, Sun Sun, Ping Nie, Kai Zhou, Ge Zhang, Ming-Hsuan Yang, and Wenhu Chen.
\newblock Context forcing: Consistent autoregressive video generation with long context.
\newblock \emph{arXiv preprint arXiv:2602.06028}, 2026.

\bibitem[Cheng et~al.(2025)Cheng, Gao, Hu, Hu, Huang, Ji, et~al.]{wananimate2025}
Gang Cheng, Xin Gao, Li Hu, Siqi Hu, Mingyang Huang, Chaonan Ji, et~al.
\newblock Wan-animate: Unified character animation and replacement with holistic replication.
\newblock \emph{arXiv preprint arXiv:2509.14055}, 2025.

\bibitem[Ephrat et~al.(2018)Ephrat, Mosseri, Lang, Dekel, Wilson, Hassidim, Freeman, and Rubinstein]{avspeech}
Ariel Ephrat, Inbar Mosseri, Oran Lang, Tali Dekel, Kevin Wilson, Avinatan Hassidim, William~T. Freeman, and Michael Rubinstein.
\newblock Looking to listen at the cocktail party: A speaker-independent audio-visual model for speech separation.
\newblock In \emph{ACM TOG}, 2018.

\bibitem[Fang and Zhao(2024)]{fang2024usp}
Jiarui Fang and Shangchun Zhao.
\newblock Usp: A unified sequence parallelism approach for long context generative ai.
\newblock \emph{arXiv preprint arXiv:2405.07719}, 2024.

\bibitem[Gan et~al.(2025)Gan, Ren, Zhang, Ye, Xie, Yin, Yuan, Peng, and Zhu]{gan2025humandit}
Qijun Gan, Yi Ren, Chen Zhang, Zhenhui Ye, Pan Xie, Xiang Yin, Zehuan Yuan, Bingyue Peng, and Jianke Zhu.
\newblock Humandit: Pose-guided diffusion transformer for long-form human motion video generation.
\newblock \emph{arXiv preprint arXiv:2502.04847}, 2025.

\bibitem[Heusel et~al.(2017)Heusel, Ramsauer, Unterthiner, Nessler, and Hochreiter]{heusel2017fid}
Martin Heusel, Hubert Ramsauer, Thomas Unterthiner, Bernhard Nessler, and Sepp Hochreiter.
\newblock Gans trained by a two time-scale update rule converge to a local nash equilibrium.
\newblock In \emph{NeurIPS}, 2017.

\bibitem[Ho et~al.(2020)Ho, Jain, and Abbeel]{ho2020denoising}
Jonathan Ho, Ajay Jain, and Pieter Abbeel.
\newblock Denoising diffusion probabilistic models.
\newblock In \emph{NeurIPS}, 2020.

\bibitem[Ho et~al.(2022)Ho, Salimans, Gritsenko, Chan, Norouzi, and Fleet]{ho2022videoddpm}
Jonathan Ho, Tim Salimans, Alexey Gritsenko, William Chan, Mohammad Norouzi, and David~J Fleet.
\newblock Video diffusion models.
\newblock \emph{arXiv preprint arXiv:2204.03458}, 2022.

\bibitem[Hu et~al.(2024)Hu, Gao, Zhang, Sun, Zhang, and Bo]{hu2024animate}
Li Hu, Xin Gao, Peng Zhang, Ke Sun, Bang Zhang, and Liefeng Bo.
\newblock Animate anyone: Consistent and controllable image-to-video synthesis for character animation.
\newblock \emph{arXiv preprint arXiv:2311.17117}, 2024.

\bibitem[Hu et~al.(2026)Hu, Gong, Yang, An, Xu, and Liu]{hu2026multianimate}
Yingcheng Hu, Haowen Gong, Chuanguang Yang, Zhulin An, Yongjun Xu, and Songhua Liu.
\newblock Multianimate: Pose-guided image animation made extensible.
\newblock \emph{arXiv preprint arXiv:2602.21581}, 2026.

\bibitem[Huang et~al.(2025)Huang, Li, He, Zhou, and Shechtman]{huang2025selfforcing}
Xun Huang, Zhengqi Li, Guande He, Mingyuan Zhou, and Eli Shechtman.
\newblock Self forcing: Bridging the train-test gap in autoregressive video diffusion.
\newblock In \emph{NeurIPS}, 2025.

\bibitem[Jafarian and Park(2021)]{tiktokdance}
Yasamin Jafarian and Hyun~Soo Park.
\newblock Learning high fidelity depths of dressed humans by watching social media dance videos.
\newblock In \emph{Proceedings of the IEEE/CVF conference on computer vision and pattern recognition}, pages 12753--12762, 2021.

\bibitem[Jiang et~al.(2025)Jiang, Han, Mao, Zhang, Pan, and Liu]{jiang2025vace}
Zeyinzi Jiang, Zhen Han, Chaojie Mao, Jingfeng Zhang, Yulin Pan, and Yu Liu.
\newblock Vace: All-in-one video creation and editing.
\newblock In \emph{ICCV}, pages 17191--17202, 2025.

\bibitem[Li et~al.(2026{\natexlab{a}})Li, Liu, Lin, and Chandraker]{li2026rollingsink}
Haodong Li, Shaoteng Liu, Zhe Lin, and Manmohan Chandraker.
\newblock Rolling sink: Bridging limited-horizon training and open-ended testing in autoregressive video diffusion.
\newblock \emph{arXiv preprint arXiv:2602.07775}, 2026{\natexlab{a}}.

\bibitem[Li et~al.(2026{\natexlab{b}})Li, Gao, Hassan, Feng, Pan, Luan, and Alahi]{li2025everanimate}
Wuyang Li, Yang Gao, Mariam Hassan, Lan Feng, Wentao Pan, Po-Chien Luan, and Alexandre Alahi.
\newblock Everanimate: Minute-scale human animation via latent flow restoration.
\newblock \emph{arXiv preprint arXiv:2605.15042}, 2026{\natexlab{b}}.

\bibitem[Li et~al.(2026{\natexlab{c}})Li, Pan, Luan, Gao, and Alahi]{li2026svi}
Wuyang Li, Wentao Pan, Po-Chien Luan, Yang Gao, and Alexandre Alahi.
\newblock Stable video infinity: Infinite-length video generation with error recycling.
\newblock In \emph{ICLR}, 2026{\natexlab{c}}.

\bibitem[Liu et~al.(2026)Liu, Hu, Xu, Shan, and Lu]{liu2026rollingforcing}
Kunhao Liu, Wenbo Hu, Jiale Xu, Ying Shan, and Shijian Lu.
\newblock Rolling forcing: Autoregressive long video diffusion in real time.
\newblock In \emph{ICLR}, 2026.

\bibitem[Peebles and Xie(2023)]{peebles2023dit}
William Peebles and Saining Xie.
\newblock Scalable diffusion models with transformers.
\newblock In \emph{ICCV}, 2023.

\bibitem[Rombach et~al.(2022)Rombach, Blattmann, Lorenz, Esser, and Ommer]{rombach2022ldm}
Robin Rombach, Andreas Blattmann, Dominik Lorenz, Patrick Esser, and Bj{\"o}rn Ommer.
\newblock High-resolution image synthesis with latent diffusion models.
\newblock In \emph{CVPR}, 2022.

\bibitem[Shi et~al.(2026)Shi, Xu, Li, Peng, Yang, Lu, Hu, and Zhang]{shi2025onetoall}
Shijun Shi, Jing Xu, Zhihang Li, Chunli Peng, Xiaoda Yang, Lijing Lu, Kai Hu, and Jiangning Zhang.
\newblock One-to-all animation: Alignment-free character animation and image pose transfer.
\newblock In \emph{CVPR}, pages 4011--4021, 2026.

\bibitem[Siarohin et~al.(2019)Siarohin, Lathuiliere, Tulyakov, Ricci, and Sebe]{siarohin2019fomm}
Aliaksandr Siarohin, Stephane Lathuiliere, Sergey Tulyakov, Elisa Ricci, and Nicu Sebe.
\newblock First order motion model for image animation.
\newblock In \emph{NeurIPS}, 2019.

\bibitem[Tong et~al.(2022)Tong, Song, Wang, and Wang]{tong2022videomae}
Zhan Tong, Yibing Song, Jue Wang, and Limin Wang.
\newblock Videomae: Masked autoencoders are data-efficient learners for self-supervised video pre-training.
\newblock In \emph{NeurIPS}, 2022.

\bibitem[Tong et~al.(2024)Tong, Li, Chen, Wu, and Zhou]{tong2024musepose}
Zhengyan Tong, Chao Li, Zhaokang Chen, Bin Wu, and Wenjiang Zhou.
\newblock Musepose: A pose-driven image-to-video framework for virtual human generation.
\newblock \url{https://github.com/TMElyralab/MusePose}, 2024.
\newblock Open-source project.

\bibitem[Wang et~al.(2018)Wang, Liu, Zhu, Liu, Tao, Kautz, and Catanzaro]{wang2018vid2vid}
Ting-Chun Wang, Ming-Yu Liu, Jun-Yan Zhu, Guilin Liu, Andrew Tao, Jan Kautz, and Bryan Catanzaro.
\newblock Video-to-video synthesis.
\newblock In \emph{NeurIPS}, 2018.

\bibitem[Wang et~al.(2025)Wang, Zhang, Tang, Zhang, Gao, Wang, and Sang]{wang2025unianimatedit}
Xiang Wang, Shiwei Zhang, Longxiang Tang, Yingya Zhang, Changxin Gao, Yuehuan Wang, and Nong Sang.
\newblock Unianimate-dit: Human image animation with large-scale video diffusion transformer.
\newblock \emph{arXiv preprint arXiv:2504.11289}, 2025.

\bibitem[Wang et~al.(2024)Wang, Li, Zeng, Fang, Guo, Liu, Tan, Chen, Xue, Dai, and Lin]{humanvid}
Zhenzhi Wang, Yixuan Li, Yanhong Zeng, Youqing Fang, Yuwei Guo, Wenran Liu, Jing Tan, Kai Chen, Tianfan Xue, Bo Dai, and Dahua Lin.
\newblock Humanvid: Demystifying training data for camera-controllable human image animation.
\newblock In \emph{NeurIPS}, 2024.

\bibitem[Williams and Zipser(1989)]{williams1989teacherforcing}
Ronald~J Williams and David Zipser.
\newblock A learning algorithm for continually running fully recurrent neural networks.
\newblock \emph{Neural Computation}, 1\penalty0 (2):\penalty0 270--280, 1989.

\bibitem[Wu et~al.(2024)Wu, Zhang, Zhang, Chen, Liao, Li, Gao, Wang, Zhang, Sun, Yan, Min, Zhai, and Lin]{qalign}
Haoning Wu, Zicheng Zhang, Weixia Zhang, Chaofeng Chen, Liang Liao, Chunyi Li, Yixuan Gao, Annan Wang, Erli Zhang, Wenxiu Sun, Qiong Yan, Xiongkuo Min, Guangtao Zhai, and Weisi Lin.
\newblock {Q-Align}: Teaching {LMM}s for visual scoring via discrete text-defined levels.
\newblock In \emph{Proceedings of the 41st International Conference on Machine Learning}, pages 54015--54029. PMLR, 2024.

\bibitem[Xiao et~al.(2024)Xiao, Tian, Chen, Han, and Lewis]{xiao2024streaming}
Guangxuan Xiao, Yuandong Tian, Beidi Chen, Song Han, and Mike Lewis.
\newblock Efficient streaming language models with attention sinks.
\newblock In \emph{ICLR}, 2024.

\bibitem[Xu et~al.(2024)Xu, Zhang, Liew, Yan, Liu, Zhang, Feng, and Shou]{xu2024magicanimate}
Zhongcong Xu, Jianfeng Zhang, Jun~Hao Liew, Hanshu Yan, Jia-Wei Liu, Chenxu Zhang, Jiashi Feng, and Mike~Zheng Shou.
\newblock Magicanimate: Temporally consistent human image animation using diffusion model.
\newblock \emph{arXiv preprint arXiv:2311.16498}, 2024.

\bibitem[Yan et~al.(2025)Yan, Ye, Yang, Teng, Dong, Wen, Gu, Liu, and Tang]{yan2025scail}
Wenhao Yan, Sheng Ye, Zhuoyi Yang, Jiayan Teng, ZhenHui Dong, Kairui Wen, Xiaotao Gu, Yong-Jin Liu, and Jie Tang.
\newblock Scail: Towards studio-grade character animation via in-context learning of 3d-consistent pose representations.
\newblock \emph{arXiv preprint arXiv:2512.05905}, 2025.

\bibitem[Yin et~al.(2024)Yin, Gharbi, Zhang, Shechtman, Durand, Freeman, and Park]{yin2024dmd}
Tianwei Yin, Micha{\"e}l Gharbi, Richard Zhang, Eli Shechtman, Fr{\'e}do Durand, William~T. Freeman, and Taesung Park.
\newblock One-step diffusion with distribution matching distillation.
\newblock In \emph{CVPR}, pages 6613--6623, 2024.

\bibitem[Yin et~al.(2025)Yin, Zhang, Zhang, Freeman, Durand, Shechtman, and Huang]{yin2025causvid}
Tianwei Yin, Qiang Zhang, Richard Zhang, William~T Freeman, Fr{\'e}do Durand, Eli Shechtman, and Xun Huang.
\newblock From slow bidirectional to fast autoregressive video diffusion models.
\newblock In \emph{CVPR}, 2025.

\bibitem[Zhang et~al.(2025)Zhang, Cao, Li, Zhao, Cui, Hou, Wu, Chen, Yu, Wang, and Ma]{zhang2025steadydancer}
Jiaming Zhang, Shengming Cao, Rui Li, Xiaotong Zhao, Yutao Cui, Xinglin Hou, Gangshan Wu, Haolan Chen, Xu Yu, Limin Wang, and Kai Ma.
\newblock Steadydancer: Harmonized and coherent human image animation with first-frame preservation.
\newblock \emph{arXiv preprint arXiv:2511.19320}, 2025.

\bibitem[Zhao et~al.(2023)Zhao, Bai, Rao, Zhou, and Lu]{zhao2023unipc}
Wenliang Zhao, Lujia Bai, Yongming Rao, Jie Zhou, and Jiwen Lu.
\newblock {UniPC}: A unified predictor-corrector framework for fast sampling of diffusion models.
\newblock In \emph{NeurIPS}, 2023.

\bibitem[Zhu et~al.(2026)Zhu, Zhao, He, Su, Li, and Zhu]{zhu2026causalforcing}
Hongzhou Zhu, Min Zhao, Guande He, Hang Su, Chongxuan Li, and Jun Zhu.
\newblock Causal forcing: Autoregressive diffusion distillation done right for high-quality real-time interactive video generation.
\newblock \emph{arXiv preprint arXiv:2602.02214}, 2026.

\bibitem[Zhu et~al.(2024)Zhu, Chen, Dai, Dong, Xu, Cao, Yao, Zhu, and Zhu]{zhu2024champ}
Shenhao Zhu, Junming~Leo Chen, Zuozhuo Dai, Zilong Dong, Yinghui Xu, Xun Cao, Yao Yao, Hao Zhu, and Siyu Zhu.
\newblock Champ: Controllable and consistent human image animation with 3d parametric guidance.
\newblock In \emph{ECCV}, 2024.

\end{thebibliography}
}

\clearpage
\setcounter{page}{1}
\maketitlesupplementary

\section{Additional Qualitative Results}
\label{sec:supp_qualitative}

\begin{figure*}[p]
    \centering
    \includegraphics[width=\linewidth]{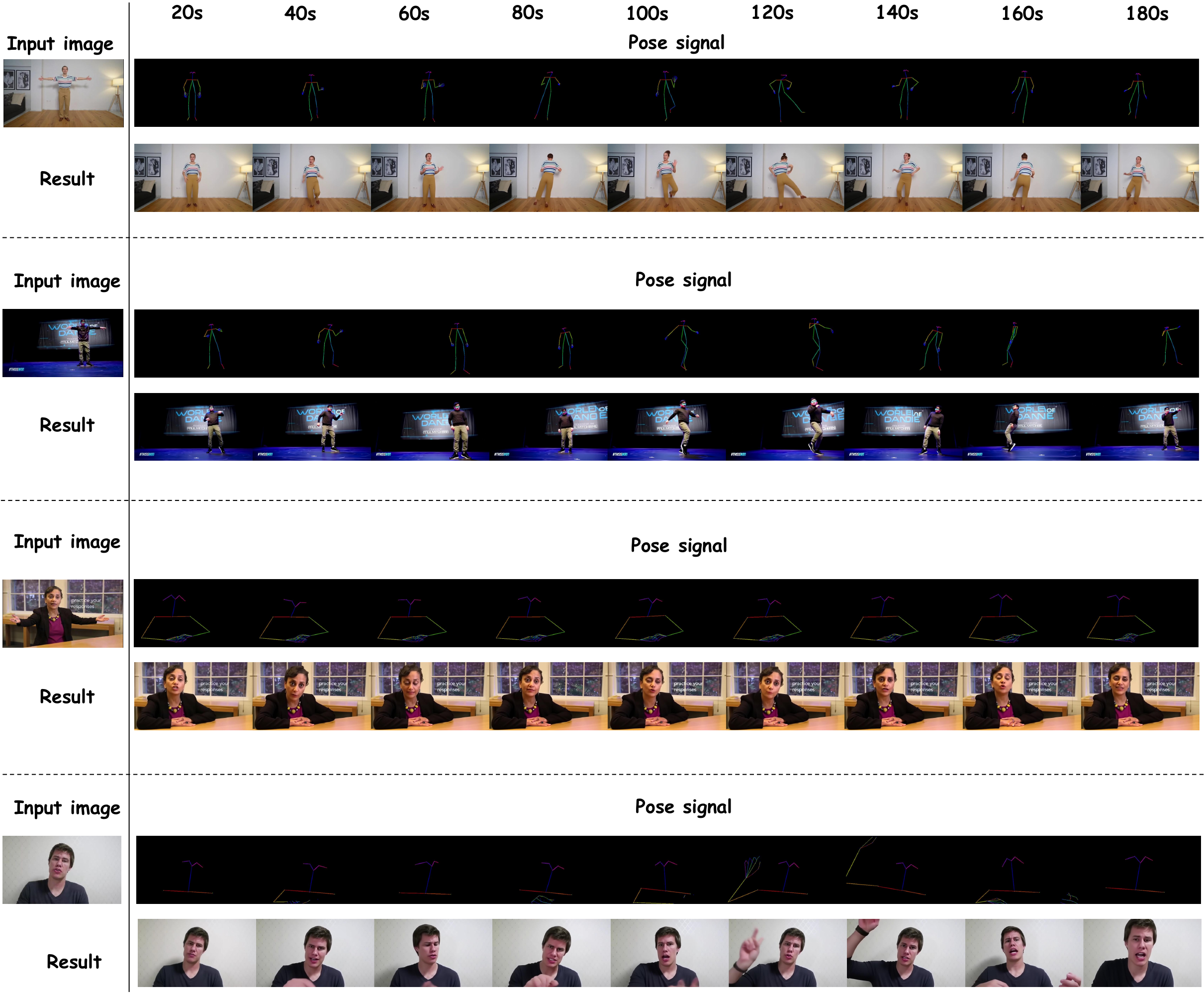}
    \caption{\textbf{Additional qualitative results on four three-minute sequences.} For each example, the left column shows the reference image, while the rows on the right show the driving pose signals and corresponding outputs sampled every 20 seconds from 20 to 180 seconds. The examples include full-body dance sequences with large pose changes and upper-body sequences dominated by facial expressions and hand motion. Across the rollout, \ours follows the input poses while preserving the reference identity, clothing, and scene appearance.}
    \label{fig:supp_qualitative_1}
\end{figure*}

\begin{figure*}[p]
    \centering
    \includegraphics[width=0.88\linewidth]{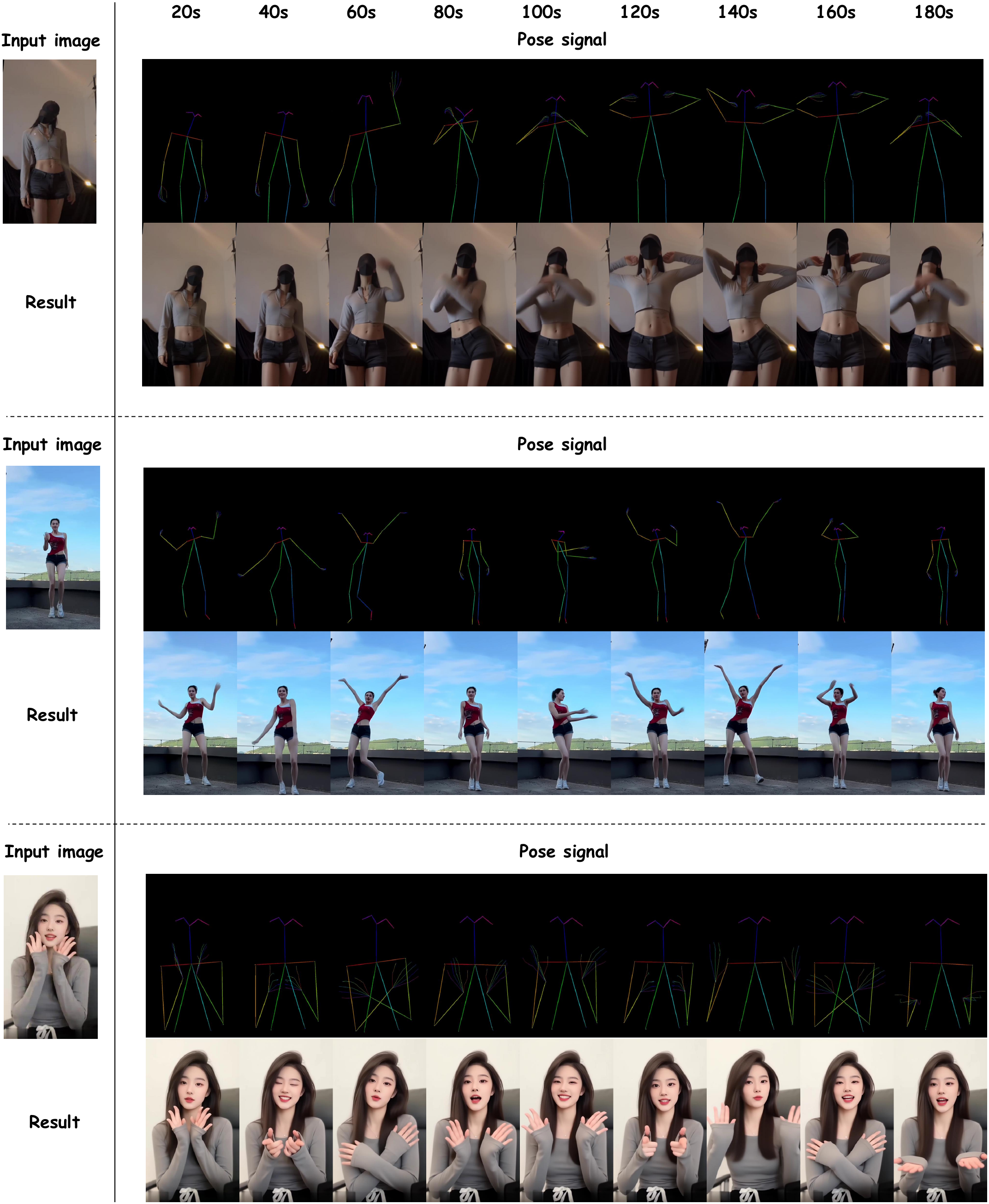}
    \caption{\textbf{Additional qualitative results on three three-minute sequences.} We show the reference image, driving pose signal, and generated outputs at 20-second intervals. The sequences cover indoor full-body motion, an outdoor dance sequence, and an upper-body subject performing frequent hand gestures. \ours maintains the subject's appearance and background while responding to substantial changes in body configuration over the three-minute rollout.}
    \label{fig:supp_qualitative_2}
\end{figure*}

We provide seven additional three-minute examples to complement the qualitative comparisons in Sec.~\ref{sec:experiments}. Figures~\ref{fig:supp_qualitative_1} and~\ref{fig:supp_qualitative_2} cover full-body and upper-body animation under diverse identities, clothing, backgrounds, camera framing, and motion patterns. For each example, we show the reference image together with the driving pose signal and generated frame at 20-second intervals from 20 to 180 seconds. These uniformly sampled results expose temporal changes throughout the rollout rather than selecting only adjacent or early frames.

The first two examples in Figure~\ref{fig:supp_qualitative_1} contain large changes in limb configuration and body orientation, whereas the last two emphasize upper-body motion, facial variation, and gestures near the face. Despite these differences, the generated subjects remain recognizable and their clothing and backgrounds do not exhibit progressive changes over the sampled timestamps. Figure~\ref{fig:supp_qualitative_2} further tests appearance preservation across a dim indoor scene, a bright outdoor scene, and a close-up portrait. The outputs track both repeated and newly observed poses while maintaining scene-specific appearance. Together with the quantitative three-minute evaluation in the main paper, these examples provide additional qualitative evidence that \ours supports stable long-form streaming across varied motion and visual conditions.

\end{document}